\documentclass[10pt,twocolumn,letterpaper]{article}

\usepackage[pagenumbers]{wacv} 

\definecolor{wacvblue}{rgb}{0.21,0.49,0.74}
\usepackage[pagebackref,breaklinks,colorlinks,allcolors=wacvblue]{hyperref}
\usepackage{multirow}
\usepackage{array}
\usepackage{float}

\def\wacvPaperID{724} 
\def\confName{WACV}
\def\confYear{2026}

\title{RoundaboutHD: High-Resolution Real-World Urban Environment Benchmark for Multi-Camera Vehicle Tracking}

\author{
Yuqiang Lin\textsuperscript{1} \quad
Sam Lockyer\textsuperscript{1} \quad
Mingxuan Sui\textsuperscript{1} \quad
Li Gan\textsuperscript{1} \quad
Florian Stanek\textsuperscript{2} \quad
Markus Zarbock\textsuperscript{2} \\
Wenbin Li\textsuperscript{1} \quad
Adrian Evans\textsuperscript{1} \quad
Nic Zhang\textsuperscript{1} \\
\textsuperscript{1}University of Bath        \textsuperscript{2}Starwit Technologies GmbH
}

\begin{document}
\maketitle
\begin{abstract}
The multi-camera vehicle tracking (MCVT) algorithms hold significant potential for various smart city applications, including anomaly detection, traffic density estimation, and suspect vehicle tracking. Developing and benchmarking such algorithms needs high quality opensource datasets. However, the publicly available datasets at the moment exhibit several limitations, such as overly simplistic scenarios, low-resolution footage, and insufficiently diverse conditions, creating a considerable gap between academic research and real-world scenarios. To fill this gap, we introduce RoundaboutHD, a comprehensive, high-resolution multi-camera vehicle tracking benchmark dataset specifically designed to represent real-world roundabout scenarios. RoundaboutHD provides a total of 40 minutes of labeled video footage captured by four non-overlapping, high-resolution (4K resolution, 15 fps) cameras. In total, 512 unique vehicle identities are annotated across different camera views, offering rich cross-camera association data. In addition to the full MCVT dataset, several subsets are also available for object detection, single camera tracking, and image-based vehicle re-identification (ReID) tasks.  Vehicle model information and camera modeling/geometry information are also included to support further analysis. We provide baseline results for vehicle detection, single-camera tracking, image-based vehicle re-identification, and multi-camera tracking. The dataset and the evaluation code are publicly available at: \url{https://github.com/siri-rouser/RoundaboutHD.git} 
\end{abstract}    
\section{Introduction}
\label{sec:intro}
With the continuous expansion of urbanization, the city traffic management is becoming more and more challenging, motivating the development of intelligent transportation systems (ITS) within the broader concept of smart cities. A practical approach to construct ITS is taking advantage of existing traffic cameras, which capture extensive amounts of useful information from video footage.

\begin{figure}
    \centering
    \includegraphics[width=1\linewidth]{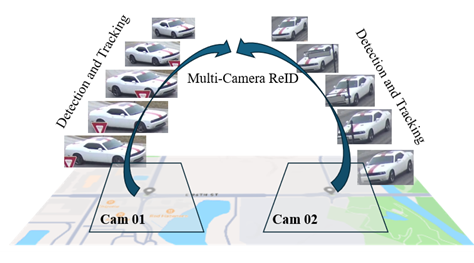}
    \caption{MCVT Problem Overview.
    The example shows two tracklet from two different non-overlapping FOV camera (Cam01 and Cam02). The cropped images above each camera represent tracklets generated from detection and single-camera tracking, and the blue arrow represents the cross-camera vehicle ReID process, linking the two tracklets as the same vehicle identity across views.}
    \label{fig:one}
\end{figure}

\begin{table*}[t]
    \centering
    \scriptsize
    \begin{tabular*}{\textwidth}{@{\extracolsep{\fill}}*{10}{c}} 
        \hline
        & \textbf{Dataset} 
        & \textbf{Cam\_Num} 
        & \textbf{Boxes} 
        & \textbf{Average Duration} 
        & \textbf{Resolution} 
        & \textbf{Fps} 
        & \textbf{Video} 
        & \textbf{Vehicle Info} 
        & \textbf{Open-Sourced} \\
        \hline
        \multirow{5}{*}{\shortstack{\textbf{Image-} \\ \textbf{based}}}
        & VehicleID \cite{liu2016pkuveri}     & /   & 221,763   & /         & /          & /   & $\times$ & $\surd$ & $\surd$ \\
        & VeRi-776 \cite{liu2016largeveri}       & 20  & 49,357    & /         & /          & /   & $\times$ & $\times$ & $\surd$ \\
        & VEVI-WILD \cite{lou2019veriwild}     & 174 & 416,314   & /         & /          & /   & $\times$ & $\times$ & $\surd$ \\
        & VRIC \cite{kanaci2019vric} & 120 & 60,430 & / & / & / & $\times$ & $\times$ & $\surd$ \\
        & Vehicle-1M \cite{guo2018learning} & / & 936,051 & / & / & $\times$ & $\surd$ & $\surd$ \\
        \hline
        \multirow{5}{*}{\shortstack{\textbf{Video-} \\ \textbf{based}}}
        & CityFlow \cite{tang2019cityflow}       & 46  & 313,931   & 4.66min   & 1280×960   & 10  & $\surd$  & $\times$ & $\times$ \\
        & SyntheVehicle \cite{herzog2023synthehicle} & 340 & 4,623,184 & 3min      & 1920×1080  & 10  & $\surd$  & $\times$ & $\surd$ \\
        & LHTV \cite{yang2022trafficlhtv}         & 16  & /         & 63.26min  & 1920×1080  & 30  & $\surd$  & $\times$ & $\surd$ \\
        & HST \cite{zhang2024videohst}          & 6   & /         & 30.33min  & /          & 30  & $\surd$  & $\times$ & $\surd$ \\
        & \textbf{RoundaboutHD (ours)} & 4   & 549,909   & 10min     & \textbf{3840×2160} & 15  & $\surd$  & $\surd$ & $\surd$ \\
        \hline
    \end{tabular*}
    \caption{Overview of Existing Vehicle-Specific MTMCT Datasets. This table provides a comparative summary of commonly used datasets for both image-based and video-based vehicle ReID. The comparison covers various aspects relevant to MTMCT tasks. A $\surd$ indicates the feature is supported, a $\times$ indicates it is not, and a / denotes that the information is unavailable.}
    \label{tab:mtmct_benchmarks}
\end{table*}

A common application of these video resources is tracking vehicles across multiple camera views, which provides vehicle trajectory information to city management authorities. This task, known as multi-target multi-camera tracking (MTMCT), has attracted considerable research interest in recent years \cite{amosa2023multi}. Generally, this MTMCT problem can be divided into 3 distinct sub-problems, as shown in \cref{fig:one}. (1) Object Detection, which detects and identifies vehicles within the camera scenes; (2) Single-Camera Tracking, which tracks vehicles’ movement across sequential frames within each camera’s field of view; and (3) Multi-Camera Re-identification, which matches and associate’s vehicle identities across separate different camera views. This identification can be performed either at an image-level (image-based re-identification) or at a tracklet-level (video-based re-identification), ultimately reconstructing complete vehicle trajectories.

Each of these sub-problems presents significant challenges in its respective domain, leading to the creation of specialized datasets for object detection, tracking, and re-identification tasks \cite{lin2014microsoft,dendorfer2020mot20,liu2016veri}. However, there is a very limited number of open-source real-world vehicle-specified MTMCT dataset has been proposed, and to the best of the authors’ knowledge, CityFlow \cite{tang2019cityflow} is currently the only open-source dataset of this kind. 

The main challenges in developing a real-world MCVT datasets arise from the small inter-class variability and large intra-class variability. Vehicles of the same model typically exhibit very subtle differences, making the re-identification process very challenging even for humans. Conversely, the same vehicle can have dramatic appearance difference due to different lighting conditions, camera viewpoint and occlusion, which further complicating the match process. Moreover, the dataset creation is also very time-consuming, requiring the annotating, labeling and linking of hundreds or even thousands of vehicles over extended periods. Considering these challenges and to enrich the available MCVT datasets, we introduce RoundaboutHD, a high resolution and challenging dataset comprising of sequences captured from four location in a small town located in US, the geographic layout is illustrated in \cref{fig:layout}. RoundaboutHD comprises in total 40 minutes of annotated video footage recorded simultaneously from four non-overlapping high-resolution cameras (4K, 15 FPS), with 10 minutes of labeled video from each camera. The dataset also introduces increased challenges, including dramatic appearance change due the change of vehicle view angle, frequent occlusions and nonlinear vehicle movements. In addition to the primary dataset designed for evaluating MCVT, we provide subsets for object detection, single camera tracking and image-based ReID. Detailed vehicle attributes (including vehicle type, color, make, and model) with comprehensive camera geometry/calibration information are also provided, enabling diverse research within this domain.

Compared to the widely used CityFlow benchmark, RoundaboutHD focus on a single scene with longer duration footage for each camera, while still maintaining a comparable number of annotated vehicle identities (512 vs. 666), demonstrating the substantial scale of the dataset. Additionally, it emphasizes real-world representativeness by offering longer, higher-resolution video, and richer contextual annotations—including vehicle attributes and camera calibration information. We hope that the release of this dataset will support and push forward research in ITS, especially for the real-world application of MCVT. The dataset, along with ground-truth annotations for detection, single-camera tracking, image-based ReID, and multi-camera tracking, as well as the associated evaluation tools, are publicly available.
\section{Vehicle ReID Datasets}
\cref{tab:mtmct_benchmarks} compares nine of the most widely used datasts for vehicle ReID. The table splits existing ReID datasets into image-based and video-based (MTMCT), covering vehicle-specified re-identification scenarios, and these scenarios are discussed in more details below.
\subsection{Image-Based Vehicle ReID Dataset}
The images-based vehicle ReID datasets \cite{liu2016veri,lou2019veriwild,liu2016largeveri,liu2016pkuveri,kanaci2019vric,guo2018learning} are designed specifically for ReID tasks at image level, excluding the detection and tracking stage. Among these datasets, VeRI-776 \cite{liu2016largeveri} is one of the most widely used benchmarks for image-based vehicle ReID. It extends the original VeRI dataset \cite{liu2016veri}, increasing the number of annotated vehicle identities from 619 to 776, with images captured from 20 camera views. A larger dataset, VERI-WILD \cite{lou2019veriwild} is notable for its large scale, comprising images collected from 174 surveillance cameras over one month in uncontrolled real-world conditions. On the other hand, the PKU-VehicleID \cite{liu2016pkuveri} stands out for its inclusion of detailed vehicle make/model annotations, offering richer vehicle identity information than many other datasets. The VRIC dataset \cite{kanaci2019vric} is built from the well-known UA-DETRAC vehicle tracking dataset \cite{wen2020uadetrac}, and aims to reflect more realistic conditions for vehicle ReID. It contains lower-resolution images from multiple viewpoints, lighting conditions and occlusion conditions, which better represents real-world vehicle appearances in wide-area traffic scenes. And finally, Vehicle-1M \cite{guo2018learning} is the largest vehicle ReID dataset published to date, it contains nearly 1 million images of 55,527 vehicles across 400 different models. All dataset mentioned above provide a solid foundation for image-based vehicle ReID research and have helped to drive the research progress toward the video-based ReID task. 
\subsection{Video-Based Vehicle ReID Dataset}
\begin{figure}
    \centering
    \includegraphics[width=1\linewidth]{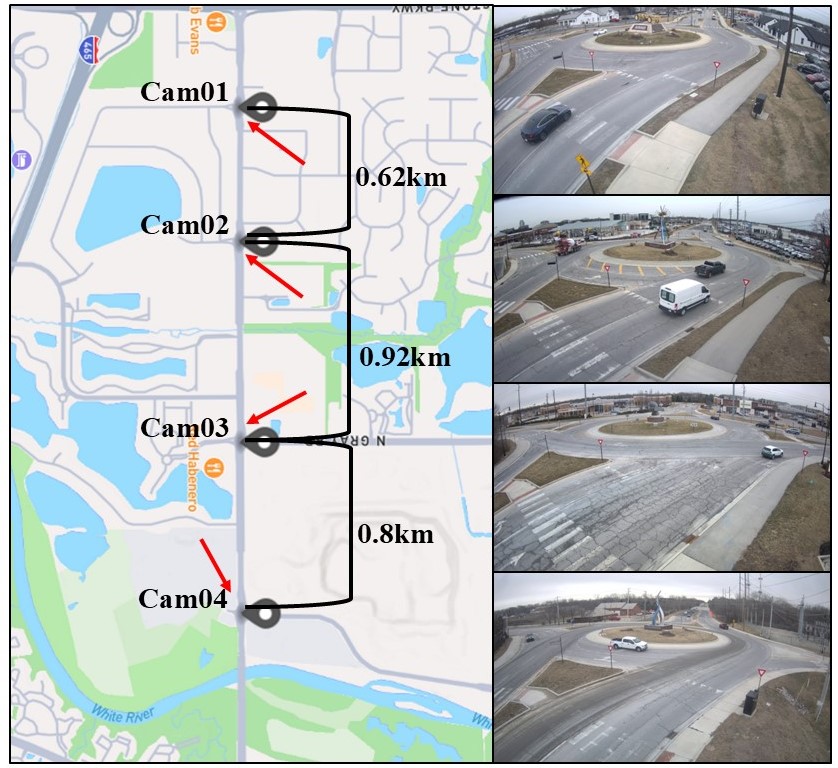}
    \caption{Camera Layout and Preview of the Proposed RoundaboutHD Dataset. Distances among camera pairs are shown, and red arrows indicate the orientation of each camera}
    \label{fig:layout}
\end{figure}
Compared to image-based ReID, video-based ReID incorporates temporal information and make it corresponds to the MTMCT task. While several MTMCT datasets have been developed for person ReID research (e.g., \cite{ristani2016duke, zheng2015peopletracking}), vehicle-specified MTMCT dataset remain scarce. To the best of the authors' knowledge, there are only two publicly available datasets work on vehicle-specified MTMCT: the CityFlow \cite{tang2019cityflow} and Synthehicle \cite{herzog2023synthehicle}. The CityFlow dataset, released through the CVPR AI City Challenge workshop \cite{Naphade21AIC21}, has been the most widely used dataset in MCVT research. It contains 3.58 hours of video from 46 cameras with 229,680 bounding boxes of 666 vehicle identities annotated. However, each camera captures only a short duration of footage, and the video quality is relatively low (960p). Given the increasing availability of high-resolution traffic cameras, these limitations reduce CityFlow’s effectiveness in representing modern real-world urban environments. The Synthehicle dataset, on the other hand, is the first synthetic dataset with identities programatically created in CARLA \cite{Dosovitskiy17carla} simulation environment which eliminates the need for manual labeling. Compared to other datasets, it offers a large amount of video content with abundant and accurate annotations, including 2D bounding boxes, 3D positions, segmentation masks, and depth information. Nevertheless, as all video content is generated in the virtual environment, this introduces a domain gap that reduces its applicability to real-world scenarios. In addition to these open-source datasets, several private video-based ReID datasets exist for MCVT evaluation, such as the Large-scale High-resolution Traffic Video (LHTV) dataset \cite{yang2022trafficlhtv} and the Highway Surveillance Traffic (HST) dataset \cite{zhang2024videohst}. Since these datasets are not publicly available, their use is high restricted in broader academic research and benchmarking.
\section{The Proposed RoundaboutHD Dataset}
\subsection{Dataset Overview}
\begin{figure}
    \centering
    \includegraphics[width=1\linewidth]{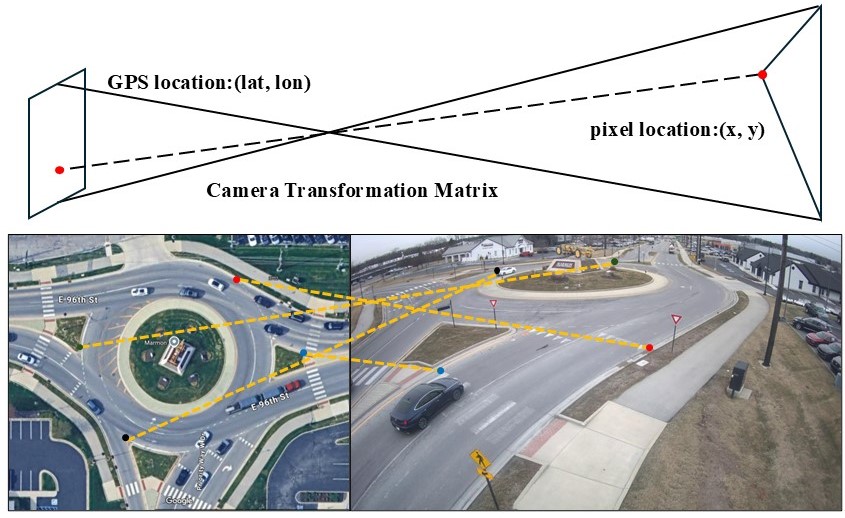}
    \caption{Illustration of the Mapping Process for Cam01. The top diagram demonstrates how camera transformation matrices are used to map pixel locations to real-world coordinates (latitude and longitude). The bottom image shows the satellite view of Cam01’s coverage area (left) and a sample video frame (right). Colored point pairs represent corresponding GPS coordinates and pixel positions, with each pair connected by a yellow dashed line.}
    \label{fig:geo-map}
\end{figure}

The proposed RoundaboutHD dataset was collected in Carmel, Indiana, a small town located in the United States with a high number of roundabouts not commonly seen in the country. The geographic layout, inter-camera distances, and sample views from each camera are illustrated in \cref{fig:layout}. The maximum distance between two cameras is approximately 0.92 km. There are several retail parks and side roads between pairs of cameras, increasing the uncertainty of vehicle movement patterns and making the tracking environment closer to real-world urban traffic scenarios.

\begin{table}[h]
    \centering
    \resizebox{\linewidth}{!}{%
    \begin{tabular}{@{}ccclclclcl@{}}
    \toprule
     & \textbf{cam01} & \multicolumn{2}{c}{\textbf{cam02}} & \multicolumn{2}{c}{\textbf{cam03}} & \multicolumn{2}{c}{\textbf{cam04}} & \textbf{Multi-Cam} \\ \midrule
    \begin{tabular}[c]{@{}c@{}}Num of \\ Vehicle\end{tabular} & 260 & \multicolumn{2}{c}{242} & \multicolumn{2}{c}{278} & \multicolumn{2}{c}{255} & 310 \\
    \bottomrule
    \end{tabular}%
    }
    \caption{Number of Annotated Vehicles in Each Camera. Multi-Cam refers to vehicles appeared in at least two cameras.}
    \label{tab:two}
\end{table}

The total duration of the RoundaboutHD dataset is 40 minutes, each 4k 15fps camera contains 10 minutes videos. All videos begin recording simultaneously to support temporally aligned multi-camera tracking. Within the 40 minutes recording, a total of 549,909 bounding boxes of 512 vehicles are manually annotated. The number of annotated vehicles per camera is summarized in \cref{tab:two}. 

\begin{figure}
    \centering
    \includegraphics[width=0.95\linewidth]{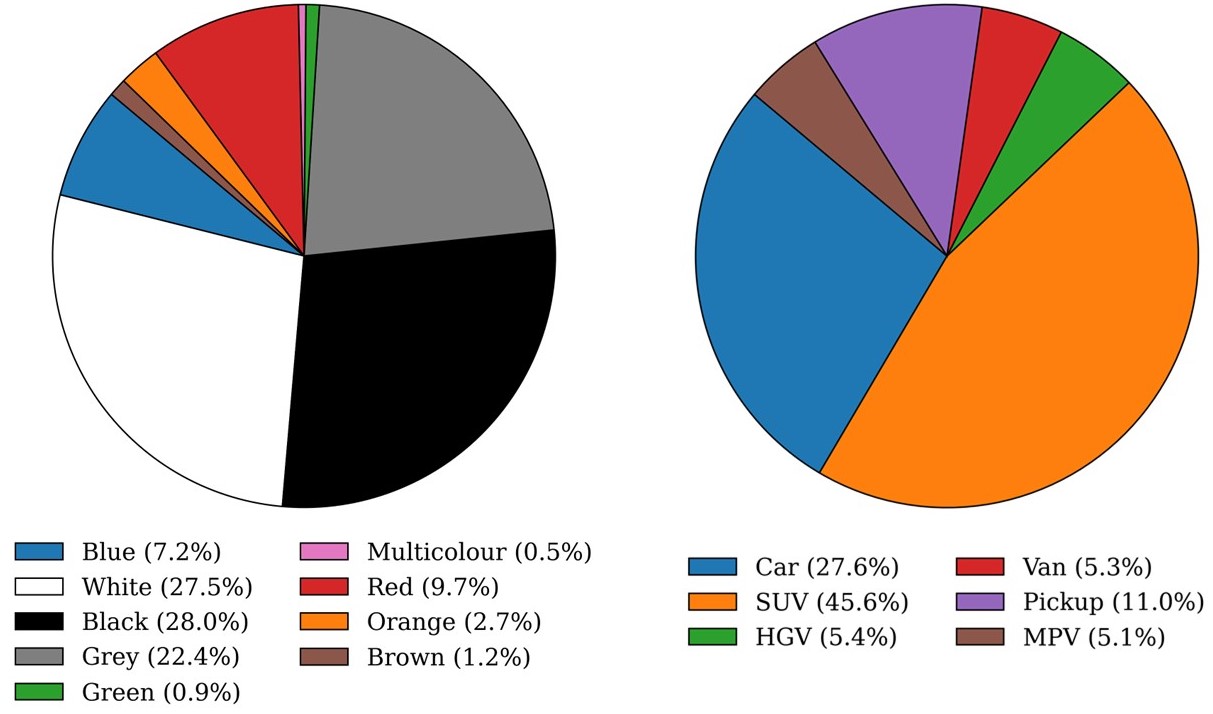}
    \caption{The Distribution of Vehicle Colors and Types in terms of Vehicle Identities in RoundaboutHD. In the vehicle type chart, \textit{Car} refers to small passenger vehicles, \textit{SUV} denotes Sport Utility Vehicles, \textit{HGV} represents Heavy Goods Vehicles, and \textit{MPV} stands for Multi-Purpose Vehicles.}
    \label{fig:veh-stat}
\end{figure}

Camera parameters are provided based on the model built in \cite{gerum2017cameratransform}, through these parameters, users can transform pixel in the image space to 3D space coordinates or to real-world GPS coordinates. An illustration of this mapping process is presented in \cref{fig:geo-map}. In addition, human annotators manually label vehicle attributes including colour, type, make, and specific model for all vehicles identified in the dataset. This rich contextual information offers valuable support for a range of future research. Based on these annotations, the five most common vehicle makes are Chevrolet, Ford, Honda, Toyota, and Nissan, while the most frequently observed vehicle models are Chevrolet Silverado, Ford Transit, Honda CR-V, Chevrolet Equinox, and Toyota Camry. A statistical summary of vehicle types and colours is provided in \cref{fig:veh-stat}.

To address the privacy concerns, a specific angle of the cameras is pre-set that ensures no recognizable human face or readable license plate is recorded. 

\subsection{Dataset Annotation}
\begin{figure}
    \centering
    \includegraphics[width=1\linewidth]{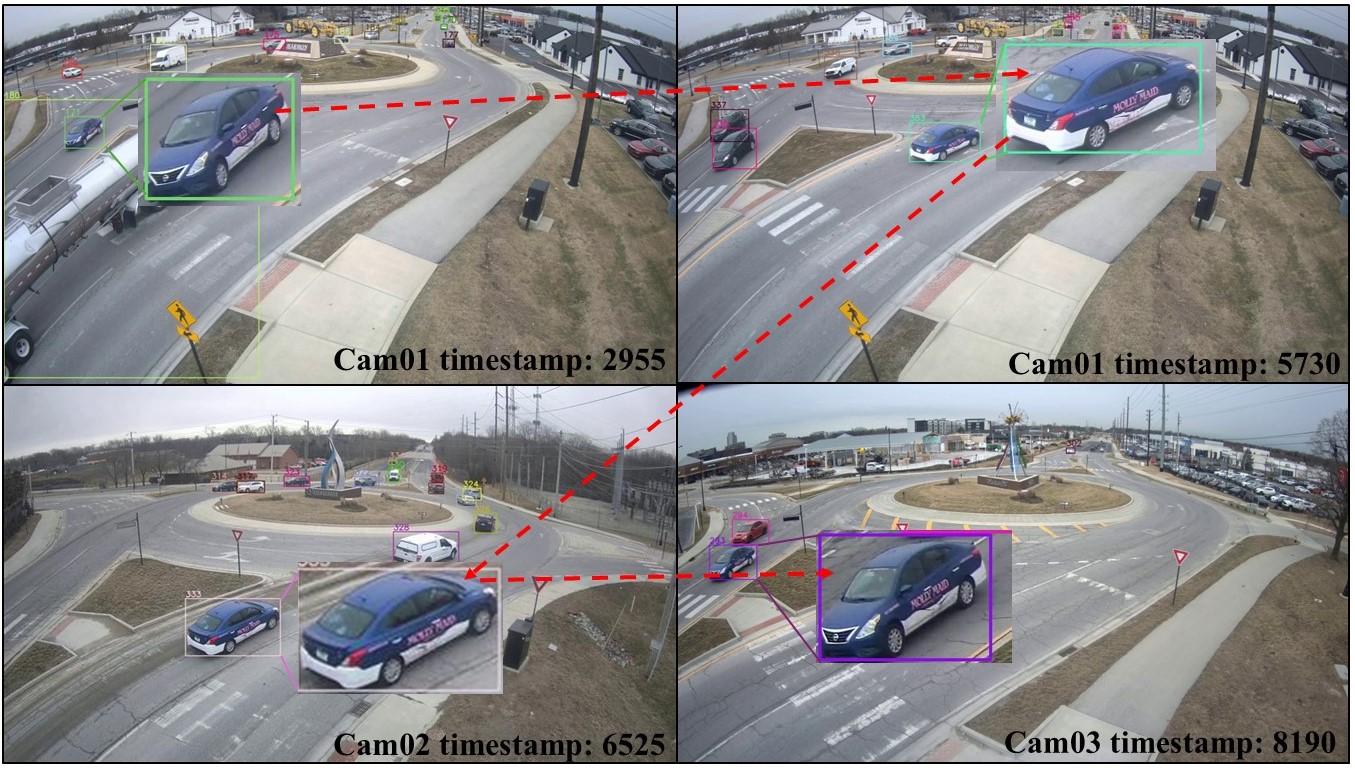}
    \caption{A Challenging Example of Cross-Camera Vehicle Trajectory Annotation. In this case, the target vehicle exits the FOV of Cam01 at timestamp 2955 and appears to head toward Cam02. However, it unexpectedly reappears in Cam01 at timestamp 5730, indicating a U-turn. This movement pattern will pose a significant challenge to MCVT algorithms that rely heavily on spatiotemporal information to constrain the search space.}
    \label{fig:annot-exam}
\end{figure}

Manual frame-level annotation for MCVT is highly time-consuming and labor-intensive. To speed up the process, we adopt a semi-automatic annotation pipeline combining detection, feature extraction, and trajectory generation with manual refinement. First, the YOLOv12x \cite{tian2025yolov12} object detector pre-trained on COCO dataset \cite{lin2014microsoft} is applied to generate initial bounding boxes. A confidence threshold of 0.35 is applied, and detections with bounding box areas smaller than 900 pixels are filtered out to reduce noise from small or distant objects. All initial detection results are manually reviewed to correct false positives and false negative identities. Next, deep appearance features are extracted using a ResNet-101 \cite{he2016deepresnet} model. These features, along with the refined detections, are inputted to SMILEtrack \cite{wang2024smiletrack} to get initial trajectories. The resulting trajectories are manually inspected and corrected to make the single-camera tracking ground truth. Based on above information, the cross-camera association is conducted manually. Vehicles are matched across different camera views based on a combination of visual attributes (color, make, and model) and spatial-temporal consistency between camera pairs. A challenging cross-camera vehicle movement example is shown in \cref{fig:annot-exam}. Finally, an image-based ReID subset is constructed based on the multi-camera association result. To ensure high-quality visual data, cropped images are extracted from bounding boxes with a minimum resolution of 3,500 pixels, sampled every 5 frames, to form the training and gallery sets. The query set is sampled every 45 frames with prioritizing high resolution cropped images. This approach ensures sufficient visual diversity and quality for robust vehicle ReID benchmarking.

\subsection{Subset Overview}
During the construction of the main MCVT dataset, we generate three task-specific subsets to support a broad range of research applications: an object detection dataset, a single-camera tracking dataset, and an image-based vehicle ReID dataset.

\textbf{Object detection dataset.} This subset includes a total of 549,909 manually verified bounding boxes, each annotated with vehicle type labels. These annotations can be directly used for training vehicle-specific object detectors, vehicle type classifiers, and other vision-based traffic applications.

\textbf{Single-camera tracking dataset.} The SCT subset includes a total of 1,042 annotated trajectories, composed of 232,945 bounding boxes. While \cref{tab:two} provides the number of unique vehicle identities per camera, the total number of annotated trajectories slightly exceeds the number of unique identities. The difference is raised from occasional long-term occlusions—often caused by large vehicles overlapping smaller ones—which leads to id switch within individual camera views. The distribution of annotated trajectories for each of the camera is summarized in \cref{tab:three}.

\begin{table}
\centering
\begin{tabular*}{\linewidth}{@{\extracolsep{\fill}}ccccc}
\hline
 & \textbf{cam01} & \textbf{cam02} & \textbf{cam03} & \textbf{cam04} \\ \hline
\begin{tabular}[c]{@{}c@{}}Num of\\ Vehicle\end{tabular} & 262 & 244 & 279 & 257 \\ \hline
\end{tabular*}%
\caption{Number of Annotated Trajectories in Each Camera.}
\label{tab:three}
\end{table}

\textbf{Image-based vehicle ReID dataset.} We construct a high-quality image-based ReID subset that is comparable in both the number of images and unique identities to several widely used ReID benchmark datasets. The dataset consists of 65,528 cropped images in total. Among these, 23,227 gallery images and 1,173 query images from 310 unique identities form the testing set. The training set includes 41,128 images from 510 unique identities. On average, each identity is represented by approximately 128.5 images, providing rich appearance variation for robust ReID evaluation.

\subsection{Benchmark Evaluation of RoundaboutHD}
To facilitate standardized benchmarking, we provide an evaluation framework for the main dataset and all subsets as part of the package. This benchmark enables comparison of different algorithms by evaluating the result text file with ground truth under task-specific performance metrics. Details on the required input file formats for each task are available in our open-source GitHub repository.

\textbf{Object detection evaluation.} For object detection, we adopt standard evaluation metrics based on Average Precision (AP), including AP@0.5, AP@0.75, AP@0.9, and the mean AP. 

\textbf{Single-camera tracking evaluation.} For SCT, we select serval metrics used by the famous MOT Challenges \cite{dendorfer2020mot20}, including IDF1 score \cite{ristani2016performance}, ID Switches and MOTA \cite{bernardin2008mota} to evaluate the performance of each SCT algorithm. 

\textbf{Image-based vehicle ReID evaluation.} For the ReID task, we follow the evaluation protocol used by the FastReID benchmark \cite{he2020fastreid}. Performance is reported using three key metrics: the mAP, rank-k accuracy, mean Inverse Negative Penalty (mINP) \cite{chen2021mINP}.

\textbf{Multi-camera vehicle tracking evaluation.} For the main MCVT dataset, we adopt the evaluation format form the well-known CVPR AICITY challenge to ensure an easy and duplicatable comparison. The evaluation framework calculates the IDF1 score \cite{ristani2016performance} (includes IDF1, IDP and IDR) as the metrics to reflect the algorithm performance. 

In the next section, the framework will be used in benchmarking experiments that helps analyzing the attributes of RoundaboutHD in comparison with existing datasets.
\section{Experiment}
In this section, we conducted several different experiments on object detection (\cref{sec:object-detection}), single camera tracking (\cref{sec:sct}), image-based ReID (\cref{sec:image-reid}), and multi-camera tracking (\cref{sec:mtmct}) using multiple SOTA methods to provide baseline performances for future research.
\subsection{Object Detection}
\label{sec:object-detection}
\begin{figure}
    \centering
    \includegraphics[width=1\linewidth]{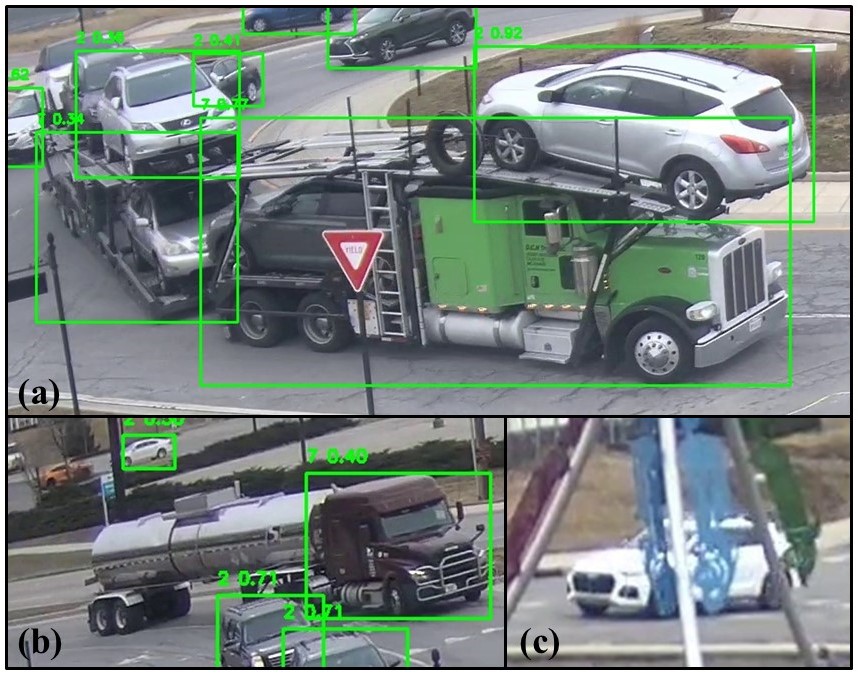}
    \caption{Example Challenging Cases for Object Detection in RoundaboutHD. (a) A combination of rare vehicle types and severe occlusion leads to both false positives and false negatives. (b) Special vehicle appearances result in incomplete detections. (c) Heavy occlusion causes false negative detection.}
    \label{fig:detection-challenge}
\end{figure}

Object detection serves as a curial part in the whole MTMCT pipeline, as its quality directly impacts the accuracy of subsequent tracking and re-identification modules. The RoundaboutHD dataset presents a variety of challenging scenarios for object detection, including occlusion, varying lighting conditions, and uncommon vehicle types. Examples of these challenges are illustrated in \cref{fig:detection-challenge}. 

Object detectors are typically classified into two categories: one-stage detectors, such as YOLO \cite{tian2025yolov12}, and two-stage detectors, such as Fast R-CNN \cite{girshick2015fastrcnn}. Since 2020, one-stage detectors—particularly the YOLO family—have become dominant in both academic research and real-world applications due to their speed and accuracy.  To evaluate how SOTA detectors perform in RoundaboutHD, we use the recently released attention-based YOLOv12x \cite{tian2025yolov12} model and transformer architecture based YOLOv11X \cite{khanam2024yolov11} as baseline model, setting the confidence threshold to 0.3, non-maximum suppression (NMS) to be true, using an inference image size of 1280, and applying an IoU threshold of 0.4. \cref{tab:yolov12} and \cref{tab:yolov11} presents its performance on the RoundaboutHD dataset, offering insights into detection performance in complex urban scenes. 
\begin{table}
    \centering
    \begin{tabular*}{\linewidth}{@{\extracolsep{\fill}}lcccc}
        \toprule
        \textbf{Camera} & \textbf{AP@0.5} & \textbf{AP@0.75} & \textbf{AP@0.9} & \textbf{mAP} \\
        \midrule
        cam01 & 40.42 & 40.25 & 37.03 & 39.23 \\
        cam02 & 69.39 & 69.22 & 64.31 & 67.64 \\
        cam03 & 74.85 & 74.70 & 71.08 & 73.54 \\
        cam04 & 79.52 & 78.90 & 74.17 & 77.53 \\
        \bottomrule
    \end{tabular*}
    \caption{YOLOv12x Object Detection Performance. All values are giving in percentage (\%).}
    \label{tab:yolov12}
\end{table}
\begin{table}
    \centering
    \begin{tabular*}{\linewidth}{@{\extracolsep{\fill}}lcccc}
        \toprule
        \textbf{Camera} & \textbf{AP@0.5} & \textbf{AP@0.75} & \textbf{AP@0.9} & \textbf{mAP} \\
        \midrule
        cam01 & 45.10 & 44.70 & 36.44 & 42.08 \\
        cam02 & 74.48 & 73.81 & 62.90 & 70.39 \\
        cam03 & 76.87 & 76.50 & 69.82 & 74.39 \\
        cam04 & 81.08 & 80.03 & 68.68 & 76.60 \\
        \bottomrule
    \end{tabular*}
    \caption{YOLOv11x Object Detection Performance. All values are giving in percentage (\%).}
    \label{tab:yolov11}
\end{table}

According to the results, the transformer-based YOLOv11x model slightly outperforms YOLOv12x model, achieving a 1.38\% higher average mAP. It is important to note that the ground truth annotations intentionally exclude static vehicles parked in the visible parking lot. However, the YOLO detector still detects these vehicles, which increases the number of false positives.  This leads to an apparent drop in detection accuracy and should be considered when interpreting the results. 

\subsection{Single Camera Tracking}
\label{sec:sct}
In the general MTMCT task, the performance of MTMCT largely depends on the performance of SCT, as errors in SCT can negatively affect downstream across cameras association. As a result, many researchers have focused on improving the SCT performance to enhance overall effectiveness of MTMCT systems. In our RoundaboutHD dataset, we provide some extra challenging for SCT. Due to the nature of roundabout, vehicles follow non-linear, curved trajectories rather than simple straight-line motion. In addition, a central statue located at the center of the roundabout (visible in \cref{fig:layout}) introduces frequent occlusions. These conditions make it more difficult to maintain consistent identity tracking over time within a single camera view.

To benchmark SCT performance on each camera of the RoundaboutHD dataset, we follow the widely used tracking-by-detection paradigm provided by BOXMOT \cite{brostrom2025boxmot}. BOXMOT is an open-source pluggable SCT framework that enables easy and straightforward deployment of multiple different SOTA SCT algorithms. In our evaluation, all trackers use the same set of raw detection results generated by the YOLOv12x model, ensuring that performance differences can be attributed solely to the tracking algorithms themselves. Additionally, to make a more representative evaluation,  trajectories corresponding to parked vehicles in visible parking areas are filtered out prior to evaluation. Detailed results from different trackers are listed from \cref{tab:boostrack} to \cref{tab:OCSort}, and a performance summary is shown in \cref{tab:SCT_summary}.
\begin{table}
    \centering
    \begin{tabular*}{\linewidth}{@{\extracolsep{\fill}}lccccc}
        \toprule
        \textbf{Camera} & \textbf{IDF1} & \textbf{IDP} & \textbf{IDR} & \textbf{IDs} & \textbf{MOTA} \\
        \midrule
        cam01 & 62.1 & 86.0 & 48.6 & 19 & 43.9 \\
        cam02 & 66.5 & 92.7 & 51.8 & 11 & 49.4 \\
        cam03 & 41.9 & 85.1 & 27.8 & 7 & 24.0 \\
        cam04 & 69.0 & 89.3 & 56.2 & 29 & 54.0 \\
        \bottomrule
    \end{tabular*}
    \caption{BoostTrack \cite{stanojevic2024boosttrack} Performance Evaluation.}
    \label{tab:boostrack}
\end{table}
\begin{table}
    \centering
    \begin{tabular*}{\linewidth}{@{\extracolsep{\fill}}lccccc}
        \toprule
        \textbf{Camera} & \textbf{IDF1} & \textbf{IDP} & \textbf{IDR} & \textbf{IDs} & \textbf{MOTA} \\
        \midrule
        cam01 & 73.9 & 75.2 & 72.6 & 127 & 80.1 \\
        cam02 & 92.2 & 95.3 & 85.7 & 9 & 83.3 \\
        cam03 & 78.5 & 80.6 & 76.5 & 94 & 83.7 \\
        cam04 & 86.1 & 89.4 & 83.0 & 36 & 83.4 \\
        \bottomrule
    \end{tabular*}
    \caption{BotSort \cite{aharon2022botsort} Performance Evaluation.}
    \label{tab:Botsort}
\end{table}
\begin{table}
    \centering
    \begin{tabular*}{\linewidth}{@{\extracolsep{\fill}}lccccc}
        \toprule
        \textbf{Camera} & \textbf{IDF1} & \textbf{IDP} & \textbf{IDR} & \textbf{IDs} & \textbf{MOTA} \\
        \midrule
        cam01 & 71.8 & 75.0 & 68.9 & 127 & 77.4 \\
        cam02 & 87.6 & 95.2 & 81.1 & 11  & 80.1 \\
        cam03 & 76.4 & 79.0 & 73.9 & 112 & 81.1 \\
        cam04 & 85.7 & 90.3 & 81.6 & 43  & 83.2 \\
        \bottomrule
    \end{tabular*}
    \caption{Deepocsort \cite{maggiolino2023deepocsort} Performance Evaluation.}
    \label{tab:Deepocsort}
\end{table}
\begin{table}
    \centering
    \begin{tabular*}{\linewidth}{@{\extracolsep{\fill}}lccccc}
        \toprule
        \textbf{Camera} & \textbf{IDF1} & \textbf{IDP} & \textbf{IDR} & \textbf{IDs} & \textbf{MOTA} \\
        \midrule
        cam01 & 72.6 & 72.2 & 73.1 & 129 & 79.0 \\
        cam02 & 90.5 & 93.3 & 87.8 & 10  & 83.7 \\
        cam03 & 84.2 & 86.9 & 81.7 & 39  & 82.4 \\
        cam04 & 86.0 & 86.8 & 85.2 & 45  & 85.3 \\
        \bottomrule
    \end{tabular*}
    \caption{ByteTrack \cite{zhang2022bytetrack} Performance Evaluation.}
    \label{tab:ByteTrack}
\end{table}
\begin{table}
    \centering
    \begin{tabular*}{\linewidth}{@{\extracolsep{\fill}}lccccc}
        \toprule
        \textbf{Camera} & \textbf{IDF1} & \textbf{IDP} & \textbf{IDR} & \textbf{IDs} & \textbf{MOTA} \\
        \midrule
        cam01 & 70.3 & 81.1 & 62.0 & 112 & 71.2 \\
        cam02 & 79.3 & 96.7 & 67.3 & 14  & 68.0 \\
        cam03 & 75.8 & 85.6 & 68.1 & 93  & 76.4 \\
        cam04 & 79.9 & 91.6 & 70.8 & 43  & 74.8 \\
        \bottomrule
    \end{tabular*}
    \caption{OCSort \cite{cao2023ocsort} Performance Evaluation.}
    \label{tab:OCSort}
\end{table}
\begin{table}
    \centering
    \begin{tabular*}{\linewidth}{@{\extracolsep{\fill}}lccccc}
        \toprule
        \textbf{Algorithm} & \textbf{IDF1} & \textbf{IDP} & \textbf{IDR} & \textbf{IDs} & \textbf{MOTA} \\
        \midrule
        BoostTrack & 59.88  & 88.28  & 46.10  & \textbf{16.5}  & 42.83 \\
        BotSort & 82.68  & 85.13  & 79.45  & 66.5   & 82.63 \\
        DeepOCSort & 80.38  & 84.88  & 76.38  & 73.25  & 80.45 \\
        \textbf{ByteTrack} & \textbf{83.33} & 84.80  & \textbf{81.95}  & 55.75  & \textbf{82.6} \\
        OCsort & 76.33  & 88.75  & 67.05  & 65.5   & 72.6  \\
        \bottomrule
    \end{tabular*}
    \caption{SCT Algorithms Performance Summary. All metrics in this table are the average values across four cameras.}
    \label{tab:SCT_summary}
\end{table}
Among all tested algorithms, ByteTrack achieves the highest average performance in terms of both IDF1 and MOTA, which are the most representative metrics for assessing overall tracking accuracy and robustness. BoostTrack reports the lowest number of ID switches; however, this might because of its limited tracking performance, as proved by its lower IDF1 and IDR score, which suggests fewer trajectories are successfully tracked. For comparison among different scenes, trackers generally perform better in Cam02 and Cam04 compared to Cam01 and Cam03. This difference is largely due to the presence of a large statue located at the center of the roundabout in Cam01 and Cam03, which blocks the view and introduces frequent occlusion during the tracking progress. 
\subsection{Image-Based ReID}
\label{sec:image-reid}
The object re-identification is an image retrieval task and represents a key component of MTMCT. It aims to match a given query image to images in a gallery, identifying the most similar gallery image that corresponds to the same object identity. The primary difference between image-based ReID and video-based ReID is that the former associates identities represented by single images, while the latter associates identities represented by multiple sequential images (tracklets).

\begin{figure}
    \centering
    \includegraphics[width=1\linewidth]{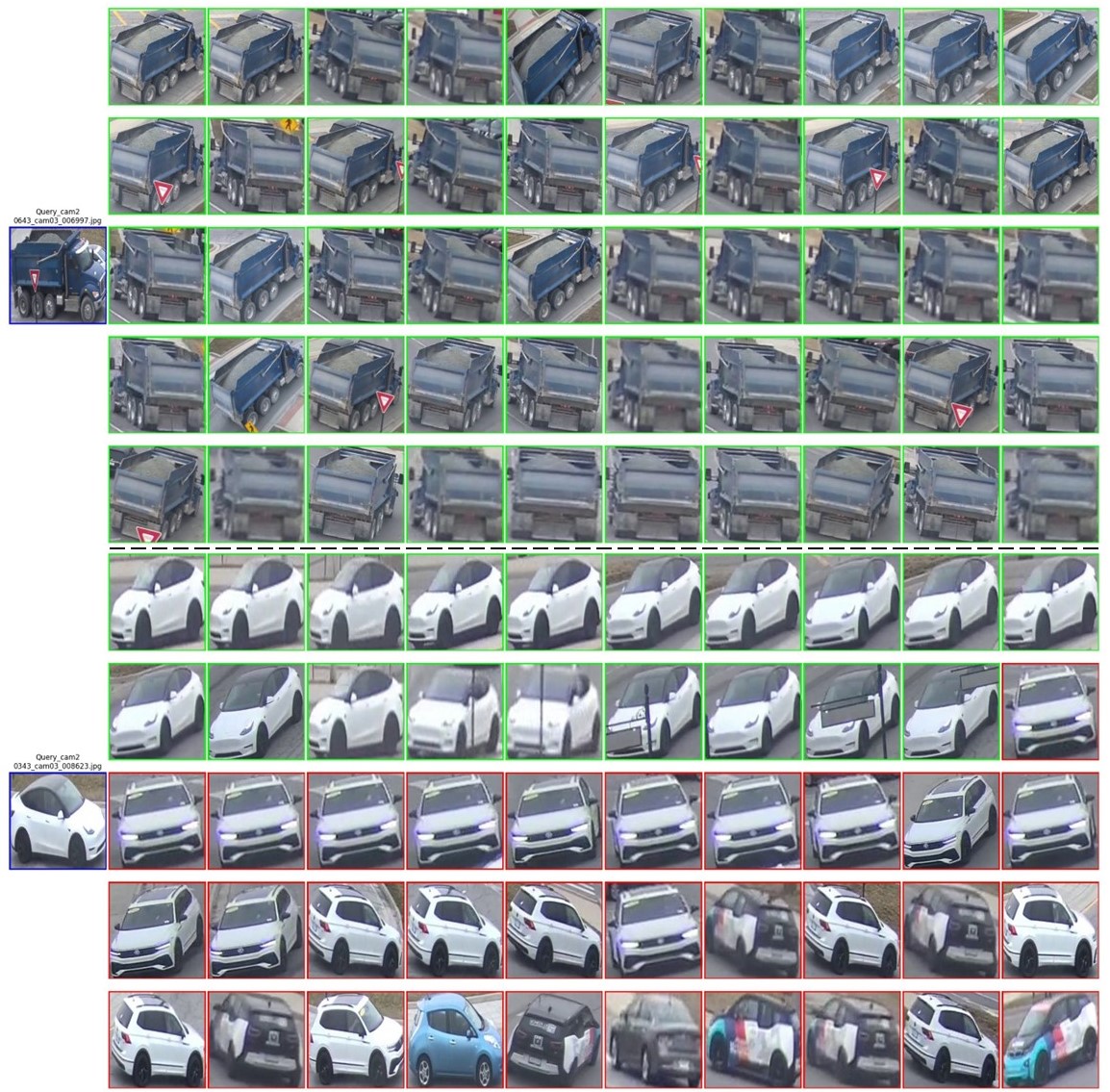}
    \caption{Rank-50 Retrieval Example for Image-Based Vehicle ReID using the SBS \cite{he2020fastreid} Model Trained on the RoundaboutHD Subset. The blue bounding box on the left marks the query image, green bounding boxes indicate correctly retrieved matches (true positives), and red bounding boxes denote incorrect matches (false negatives).}
    \label{fig:reid-example}
\end{figure}

To benchmark the RoundaboutHD image-based ReID subset, we evaluate three commonly used baseline models: BoT \cite{luo2019bot}, AGW \cite{ye2024agw}, and SBS (a stronger baseline built upon BoT) \cite{he2020fastreid}, all implemented using the FastReID framework \cite{he2020fastreid} with a shared ResNet50-IBN backbone. Each model is evaluated under two settings: (1) directly using ImageNet-pretrained weights \cite{deng2009imagenet}, and (2) after fine-tuning on our custom RoundaboutHD training set. For the fine-tuning process, we follow FastReID’s default training configuration, using a combination of CrossEntropyLoss and TripletLoss. Models are trained for 60 epochs, and the best-performing checkpoint based on validation accuracy is selected for final evaluation. All training is conducted on an NVIDIA GeForce RTX 4070 Ti GPU. The quantitative results for both settings are presented in \cref{tab:reid-no-pre-training} and \cref{tab:reid-custom}, while \cref{fig:reid-example} illustrates two visual examples from the ReID benchmark.
\begin{table}
    \centering
    \begin{tabular*}{\linewidth}{@{\extracolsep{\fill}}lcccc}
        \toprule
        \textbf{Algorithm} & \textbf{mAP} & \textbf{Rank-1} & \textbf{Rank-5} & \textbf{mINP} \\
        \midrule
        BoT \cite{luo2019bot}   & 3.08  & 15.13 & 25.71 & 0.42 \\
        AGW \cite{ye2024agw}    & 2.71  & 8.94  & 16.68 & 0.60 \\
        SBS \cite{he2020fastreid} & 19.51 & 44.80 & 58.73 & 3.56 \\
        \bottomrule
    \end{tabular*}
    \caption{Vehicle ReID performance. This table shows the results using models pre-trained only on ImageNet; all values are percentages.}
    \label{tab:reid-no-pre-training}
\end{table}
\begin{table}
    \centering
    \begin{tabular*}{\linewidth}{@{\extracolsep{\fill}}lcccc}
        \toprule
        \textbf{Algorithm} & \textbf{mAP} & \textbf{Rank-1} & \textbf{Rank-5} & \textbf{mINP} \\
        \midrule
        BoT \cite{luo2019bot}   & 98.45 & 98.73 & 99.05 & 98.41 \\
        AGW \cite{ye2024agw}    & 97.86 & 97.76 & 99.05 & 93.74 \\
        SBS \cite{he2020fastreid} & 99.19 & 99.66 & 99.66 & 98.57 \\
        \bottomrule
    \end{tabular*}
    \caption{Vehicle ReID performance. This table shows the performance of models re-trained on our custom dataset, all values are in percentage.}
    \label{tab:reid-custom}
\end{table}

According to the benchmarking results, \cref{tab:reid-no-pre-training} reports the performance of the three baseline models using only ImageNet-pretrained weights, while \cref{tab:reid-custom} presents the results after re-training on the RoundaboutHD ReID subset. Comparing \cref{tab:reid-no-pre-training} and \cref{tab:reid-custom}, the re-trained model shows an obvious advantage in terms of our selected metrics. This result highlights the domain gap between generic image datasets and real-world vehicle ReID tasks. Meanwhile, the huge performance increment also proves the effective and comprehensive of the RoundaboutHD ReID subset, which provides vehicle images shot from diverse angles and facilitates more robust feature learning for image-based ReID tasks. However, even with strong overall performance on re-trained model which reaches nearly perfect scores in mAP, Rank-5, and mINP, challenges still remain at larger retrieval depths. As shown in \cref{fig:reid-example}, the top example demonstrates excellent retrieval of a blue Mack truck, with 100\% AP on the rank-50 result. In contrast, the bottom example shows a limited performance for a white Tesla Model Y on rank 19th to rank 50th. This suggests the model finds rare or visually distinctive vehicles more easily than common ones. This is also consistent with the vehicle distribution in \cref{fig:veh-stat}, where only 7.2\% of vehicles are blue and 5.4\% are HGVs, compared to 27.6\% white vehicles and 45.6\% SUVs, which are more common and harder to distinguish.

\subsection{Multi-Camera Tracking}
\label{sec:mtmct}
Multi-camera tracking (MCT) is the task for associating vehicle identities across different camera views. In most of studies, MCT builds upon the outputs of SCT, associating identities by comparing distances between deep embedding features. Some approaches further incorporate camera topology and spatiotemporal reasoning—using inter-camera distances to estimate transition probabilities and narrow down the searching space for more accurate cross-camera association.

Following the release of CityFlow and the CVPR AI City Challenge workshop, there has been significant several efforts focusing on the MCVT problem and achieving good results. Notable examples include ELECTRICITY \cite{qian2020electricity}, the winning method of the 2020 AI City Challenge, Liu \etal. \cite{liu2021citywin}, winner in 2021, and Yang \etal. \cite{yang2022citywin}, winner in 2022. In our benchmark evaluation, we choose ELECTRICITY as the baseline method due to its general and reproducible pipeline. Unlike Liu and Yang’s approaches, which rely on manually annotated spatial-temporal zone-based information between cameras that are only available for the CityFlow dataset, ELECTRICITY can be more broadly applied to other datasets without requiring prior human annotations. During the evaluation on RoundaboutHD, we filter out all static vehicle trajectories to remove the influence of parked vehicles visible in the scene. For the cross-camera matching stage, we fine-tune the threshold distance to 12 and set a hard removal distance of 80. The performance of ELECTRICITY across multiple datasets is summarized in \cref{tab:mcmt-eval}.

\begin{table}
    \centering
    \begin{tabular*}{\linewidth}{@{\extracolsep{\fill}}lccc}
        \toprule
        \textbf{Dataset} & \textbf{IDF1} & \textbf{IDP} & \textbf{IDR} \\
        \midrule
        CityFlow & 46.16 & / & / \\
        Synthehicle & 41.50 & 48.0 & 37.5 \\
        \textbf{RoundaboutHD (Ours)} & 28.14 & 26.45 & 30.06 \\
        \bottomrule
    \end{tabular*}
    \caption{ELECTRICITY Performance Evaluation on Different Datasets.}
    \label{tab:mcmt-eval}
\end{table}

This result provides a baseline of how SOTA algorithm performances in different MCVT datasets, with IDF1 score serving as the primary metric. Moreover, it also reinforces the increased challenge posed by RoundaboutHD compared to existing dataset, which is largely attributed to nonlinear vehicle trajectories and frequent occlusions inherent in roundabout environments. 
\section{Conclusion}
This paper presents RoundaboutHD, a comprehensive, high-resolution real-world dataset designed to advance research in MCVT. Unlike previous open-source datasets, RoundaboutHD addresses the lack of high-quality dataset that closely reflect real-world conditions, and it also introduces additional challenges such as nonlinear vehicle motion and frequent occlusions, which arise naturally from the roundabout scenario.

The dataset comprises video footage from four non-overlapping FOV cameras, each offering 10 minutes of fully annotated 4K, 15 fps video. In total, it includes 549,909 annotated bounding boxes across 512 unique vehicle identities. In addition to the full MCVT dataset, we also provide subsets for object detection, single-camera tracking, and image-based vehicle ReID. The dataset further includes camera modeling/geometry and vehicle-level contextual information(vehicle type, color, make and model), which enriches its applicability to multi-modal and multi-task scenarios. To support standardized evaluation, we provide a unified evaluation framework along with baseline results from several SOTA algorithms across all tasks. These experimental results prove that RoundaboutHD presents realistic and substantial challenges, making it a valuable dataset for the evaluation of future MCVT methods.

Despite its strengths, the focus on a single scene may limit the dataset scenario diversity, which may affect model generalization. To address this, future extensions will aim to incorporate a boarder range of conditions, including different lighting, weather, and traffic patterns. We hope that the release of RoundaboutHD will inspire further advancements in the MTMCT research community.

{
    \small
    \bibliographystyle{ieeenat_fullname}
    \bibliography{main}
}
\end{document}